# A Reusable AI-Enabled Defect Detection System for Railway Using Ensembled CNN

Rahatara Ferdousi[1*], Fedwa Laamarti[2,1], Chunsheng Yang[3], Abdulmotaleb El Saddik[2,1]

[1*]School of Electrical Engineering and Computer Science, University of Ottawa, 800 King Edward, Ottawa, K1N 6N5, ON, Canada. [2]Department, Mohamed bin Zayed University of Artificial Intelligence, Masdar City, 10587, Abu Dhabi, UAE. [3]National Research Council,, 1200 Montreal Rd, Ottawa, K1A 0R6, ON, Canaada.

*Corresponding author(s). E-mail(s): rferd068@uottawa.ca;
Contributing authors: flaamart@uottawa.ca; chunsheng.yang@nrc.gc.ca ; elsaddik@uottawa.ca;

## Abstract

Accurate Defect detection is crucial for ensuring the trustworthiness of intelli-gent railway systems. Current approaches rely on single deep-learning models, like CNNs, which employ a large amount of data to capture underlying patterns. Training a new defect classifier with limited samples often leads to overfitting and poor performance on unseen images. To address this, researchers have advo-cated transfer learning and fine-tuning the pre-trained models. However, using a single backbone network in transfer learning still may cause bottleneck issues and inconsistent performance if it is not suitable for a specific problem domain. To overcome these challenges, we propose a reusable AI-enabled defect detec-tion approach. By combining ensemble learning with transfer learning models (VGG-19, MobileNetV3, and ResNet-50), we improved the classification accu-racy and achieved consistent performance at a certain phase of training. Our empirical analysis demonstrates better and consistent performance compared to other state-of-the-art approaches. The consistency substantiates the reusability of the defect detection system for newly evolved defected rail parts. Therefore we anticipate these findings to benefit further research and development of reusable AI-enabled solutions for railway systems.

# 1 Introduction

The issue of railway defect detection has become increasingly complex, as the defect domain in railways rapidly changes. For instance, some defects may closely resem-ble normal railway components, making it difficult to visually differentiate them and emphasizing the need for reusable detection methods. Even with progress in Com-puter Vision (CV) techniques, Railway Defect detection is challenging due to multiple factors such as the complexity of railway infrastructure, differences in lighting and weather conditions, and the inclusion of noise in image data. With the emergence of Artificial Intelligence technology, more advanced CV techniques are being required for monitoring and analysis of railway infrastructure [1].In railway, there is a need to develop effective methods for managing defect detection which can classify defects consistently.

Numerous studies have already been conducted for Railway defect detection, mak-ing use of various CV techniques [2–4]. However, there are some notable limitations including reliance on handcrafted features and difficulty handling complex images. With the recent development of AI models, integration of Deep Learning (DL) models into the Digital Twin (DT) framework has shown promising results in railway defect Detection [5]. The employment of DL has been increasing when it comes to auto-mated defect detection within railway infrastructure [2] and [3].In addition, the use of Convolutional Neural Network (CNN) based DL models has become popular due to their ability to detect various types of railway defects with high accuracy. Neverthe-less, there are still gaps that need to be addressed in this field. The requirement for a large sample size for training a new detection classifier is one of them.

Training a deep learning model from scratch with fewer samples can sometimes lead to lower accuracy [6]. Transfer learning has been commonly used to address the cost and complexity issues of training a CNN model with huge data for a long training period [4]. Fine-tuning pre-trained CNN allow new models to utilize the knowledge extracted from large-scale datasets by transferingr the knowledge [7]. However, in the context of railway defect detection, using single backbone network for transfer learning can introduce challenges and inconsistencies due to the dynamically changing nature of defect patterns and variations in railway infrastructure.

In our previous work we introduced a conceptual DT framework called RailTwin for railway maintenance [8]. In this research, we focus on detailing the AI Inferencing engine of this framework for defect detection of rail freights. Our main contributions of this paper are as follows:

1. We propose the integration of ensembled and fine-tuned models for reusable AI-enabled railway defect detection. Although this approach has been applied in other domains, such as medical imaging analysis [9], its application to railway defect detection is new (to the best of our knowledge).
2. We designed an algorithm for ensembling fine-tuned CNN models that optimize the trade-off between validation loss and training loss.
3. We conducted a thorough complexity analysis of the proposed algorithm. This analysis provides insights into the performance characteristics of the algorithm for further implementation in different contexts.

We conducted three distinct experiments utilizing the dataset from the Canadian Pacific Railways (CPR) dataset, which includes 1000 defective and 2000 normal rail parts images of real heavy freights.
4. We found data augmentation technique improves the accuracy but still shows significant overfitting issues for the CPR dataset. We identified that identical defective and normal images exist in CPR data, which poses a challenge to generalizing a CNN model.
5. We empirically evaluated the impact of transfer learning with VGG-19, MobileNetV3, and ResNet50 models on CPR dataset. This approach effectively reduces overfitting and ensures high accuracy. Among these models, ResNet achieves the highest accuracy, while MobileNet exhibits the least loss.
6. We implemented and evaluated our proposed model, which achieves 99% predic-tion accuracy, reduces overfitting, and maintains stable performance after a certain epoch. By comparing our results with existing work, we demonstrate that our model outperforms other state-of-the-art models in classifying defects.

   The remainder of the paper is organized as followings in Section 2, we outlined how DL has been used for digital twin defect detection in railways. Also, we summarized the data employed for this purpose.The related work is followed by Section 3, where we detailed the process of the proposed reusable AI model as well as the rationale for selecting transfer learning and ensembling as the underlying technology for rail-defect detection. In Section 5, we presented the data, experiment, and outcomes of the experiment with a description. Finally, we concluded the research by outlining the limitations and viable future extension of this study.

## 2 Related Work

The detection of railway defects is an essential task in the railway industry to ensure the safe and efficient operation of trains. The traditional inspection method is labor-intensive and time-consuming, hence the need for advanced techniques to automate this process. CV-enabled technologies have been proposed as viable solutions for rail-way defect detection [8]. In this section, we present related work on CV methods of defect detection in railways.

### 2.1 Supervised Machine Learning

Supervised machine learning (SVM) algorithms have been applied to various components of the railway system. Authors in [10], used SVM for classifying tracks in grinding condition and tracks with severe damage with an accuracy of over 95%. How-ever, SVM can be particularly useful when the data is well-structured and the features are well-defined, as it can provide a simple and efficient way to separate the classes and classify new samples [11].

### 2.2 Deep Learning

Deep learning algorithms, such as CNN [12], has significantly improved the accuracy and efficiency of defect detection. CNNs are a class of deep artificial neural networks that rely on local linear operations followed by non-linear transformations, creating different representations of the input data [13]. They have been shown to extract low-level features such as object edges and high-level features such as object shapes, considering the spatial context [14].

Table 1: Defect Detection in Railway using CV

| Study | Method | Models Used | Defects Detected | Accuracy |
|---|---|---|---|---|
| [10] | Image classification | SVM | Tracks in grinding condition, tracks with severe damage | >95% |
| [15] | Image Classification | Deep CNN (DCNN) | Normal rail, small defects, squats | 92% |
| [16] | Image Classification classification | DCNN | Normal rail, trivial defects (seed squats), squats | 96.9% |
| [17] | Object detection | YOLOv3 | Defects in rails | 97% |
| [17] | Object detection | MobileNet +YOLO | Three types of rail surface defects | 87.40% |
| [18] | One-dimensional CNN | 1D CNN | Fasteners in different degradation conditions | High detection accuracy |
| [19] | Freight defect | Image Classification | Deep CNN | Recall-80.48%, Precision-78.20%, F1-score-79.32% |
| [20] | Semantic segmentation | YOLO | Freight defects | 57.4% |
| [20] | Object detection | Rail Surface | MobileNet and YOLO | 96.3% |
| [21] | Semantic segmentation | Modified AlexNet, VGG | Rail Surface | 90% |
| [22] | Rail inspection | ResNet50 transfer learning | Surface defects | 94% |
| [23] | Image segmentation and classification | RESNET and DenseNET | Rail Track | 90% |
| [24] | Sequence data classification | 1DCNN and long- and short-term memory (LSTM) | Rail Surface | Dataset-1: Recall 0.9314, Precision 0.8421, F1-Measure 0.8845; Dataset-2: Recall-0.9427, Precision-0.9176, F1-score 0.9300 |
| [25] | Transfer Learning | Modified YOLOV4 | Faulty Valves | 92.3%-96.3% |
| [26] | Transfer Learning + Object Detection | VGG + YOLO | Fasteners | 97.28% |
| [27] | Image classification, Transfer learning | Squeezenet, Googlenet, Inception, Densenet, Mobilenet, Resnet Xception, Efficientnet | Squats | 81.6%-91.89% |

Based on the Table 1, it is evident that CNN has been widely employed for railway defect detection, using various methods such as image classification, object detection, semantic segmentation, and sequence data classification. The reason behind the popularity of using CNN for defect classification is- its ability to process and learn from large amounts of complex data, such as images and sensor readings, and to detect unseen patterns (e.g., crack) and anomalies (e.g., faulty valves) that may be difficult to discern using traditional machine learning or rule-based methods [28].

The authors in [15], used a Deep CNN model to detect normal rails, small defects, and squats, achieving an accuracy of 92%. Similarly, the evaluation in [16], demon-strated high accuracy at 96.9% accuracy in detecting normal and defective squats. Authors in [18], used one-dimensional CNN to detect fasteners in different degrada-tion conditions with high accuracy. The research in [19], used Deep CNN for image classification of freight defects, achieving a recall of 80.48%, precision of 78.20%, and F1-score of 79.32%. The study [17], applied object detection with YOLOv3 to detect defects in rails with a high accuracy of 97%. Another study in [29], used MobileNetv2 and YOLOv3 for object detection, achieving 87.4% accuracy in detecting three types of rail surface defects.

## 2.3 Semantic Segmentation

Some studies have also applied semantic segmentation-based object detection, which involves dividing an image into segments and assigning each segment a label, for defect detection. For example, Liang et al.[20], used YOLO for semantic segmentation of freight defects, achieving an accuracy of 57.4%. Authors in [21], used modified AlexNet and VGG for semantic segmentation of rail surfaces, achieving an accuracy of 90%. Semantic segmentation is applied for defect classification due to its ability to identify defects that occur within specific regions or structures in the image, such as cracks or breaks in a rail surface or joints between rail segments [30].

## 2.4 Transfer Learning

Transfer learning, a technique where a pre-trained model is used as a starting point for a new task, has also been applied to defect detection in railway components [9]. Transfer learning has been useful in railway defect detection by leveraging pre-trained models to fine-tune on a smaller dataset [31]. However, this approach reduces the risk of overfitting and allows the model to learn features from a larger, more diverse dataset. Santur et al. [22] used ResNet50 transfer learning for rail inspection, achieving 94% accuracy in detecting surface defects. The authors in [27] used transfer learning with ten CNN architectures (e.g., VGG-19, ResNet50, GoogleNet, etc.) for squat detection, achieving an accuracy range of 81.6%-91.89%. Anwar et al.[25], used modified YOLOv4 with transfer learning for the detection of faulty valves, achieving an accuracy range of 92.3%-96.3%. Chen et al. [26], used VGG and YOLO with transfer learning for fastener detection, achieving an accuracy of 97.28%.

## 2.5 Fine Tuning

Fine-tuning can also be useful in railway defect detection for adjusting pre-trained models to better fit specific railway datasets. For example, a pre-trained CNN model can be fine-tuned on a dataset of railway images containing different types of defects to learn the specific patterns and features of each defect. By fine-tuning a pre-trained model, it can be customized to better fit the characteristics of the specific dataset and improve its accuracy in detecting railway defect[32]. Transfer learning involves keeping the pre-trained weights and modifying only the final layers, while fine-tuning involves modifying and retraining the entire model on a new dataset.

## 2.6 Ensemble CNN

Ensembled CNN can also be used in railway defect detection by combining the outputs of multiple CNN models. For example, an ensemble of CNN models can be used to detect different types of railway defects such as rail cracks, joint defects, and wheel defects. Each CNN model can be trained on a specific subset of railway defect images, and the ensemble of models can provide more accurate and robust predictions by combining the strengths of each individual model [33].

## 2.7 Summary

Overall state of the art in railway defect inspection using data is diverse and mainly relies on CNN-based techniques. Different data sources and data sizes are used to train and validate models, including camera-mounted vehicles, line scan cameras, and recordings over a track. Data preprocessing techniques such as resizing, noise reduc-tion, segmentation, and data augmentation have been commonly used to improve the quality and quantity of the data. However, there are following issues in the existing literature still to be addressed.

- Backpropagation in the traditional approach is complex and time-consuming as it depends on a trial and error approach to determine the appropriate parameters.
- There is still a scarcity of suitable large datasets for heavy freight defect classifi-cation. Training a defect classifier with new samples often suffers from overfitting issues and affects the consistency of the classifier's performance
- Although transfer learning and fine-tuning are good candidate solutions for sub-duing the cost and complexity of backpropagation with limited data, the use of a single backbone network may cause bottleneck issues in case it is not suitable for a specific classification problem and unable to handle unseen images.

# 3 Methodology

In this section, we detail the architecture for the proposed reusable defect detection in railway. We choose Transfer learning and Ensemble Learning as the fundamental process of the proposed architecture because developing an intelligent defect classifier requires an accurate, fast, and esily modifiable model [14]. In this context, the inte-gration of fine-tuned CNNs could further improve the efficiency of defect detection systems. In particular, the aim of the proposed approach is to alleviate the burden of selecting all parameters from scratch to train efficient CNN models [34] as well as to ensure consistent performance for new and moderate sample sizes.

## 3.1 Problem Formulation

Let us consider D as the dataset containing rail and wheel defect images. Ini-tially, we need to train a set of n fine-tuned CNN models, denoted as M = {$CNN_1$, $CNN_2$, . . . , $CNN_n$}, using transfer learning.

Each CNN model $CNN_i$ is initialized with pre-trained weights $w_i$ obtained from pre-trained CNNs, such as VGG19, RESNET50, and MobileNetV2. The fine-tuning process adapts the models to the defect detection task using the dataset D, optimizing the model parameters $\theta_i$ to minimize the loss function $L_i$:

$$\theta_i^* = \arg\min_{\theta_i} L_i(D, w_i, \theta_i) \quad (1)$$

Then, the predictions of the fine-tuned models are combined using an ensembling function f(w), which calculates the weighted average of the individual model predictions:

$$\hat{y} = f(w_1, w_2, \ldots, w_n) \quad (2)$$

To give more weight to models with lower loss values, the weights $w_i^*$ are selected based on the individual loss values $L_i$:

$$w_i^* = \frac{1}{L_i}, \quad \text{for } i = 1, 2, \ldots, n \quad (3)$$

where the relationship between the weight is reciprocal to the loss value.

After the optimal weights $w_i^*$ that maximize the accuracy of the ensemble model $M_e$ while minimizing the loss is found:

$$w^* = \arg\max_{w_i} \text{Acc}(f(w_1, w_2, \ldots, w_n)) - \lambda L_{val} \quad (4)$$

where Acc(·) represents the accuracy metric, and $\lambda$ is a parameter controlling the trade-off between accuracy and loss.

After training the ensembled fine-tuned CNN model $f_{ens}$ for a certain number of epochs, we evaluate the performance of the models on a validation set. The objective is to minimize the validation loss while ensuring its consistency over epochs., the optimization problem can be represented as Equation (5).

$$\min_{\theta} L_{val}(\theta) \quad (5)$$

subject to

$$|L_{val}(\theta_t) - L_{val}(\theta_{t+1})| \leq \epsilon \quad (6)$$

where $L_{train}$ denotes the training loss and $L_{val}$ denotes the validation loss. $\theta$ represents the model parameters, $\theta_t$ and $\theta_{t+1}$ are the model parameters at consecutive epochs, and $\epsilon$ is a threshold that determines the acceptable change in the validation loss. The closer the value of $\epsilon$ is to 0 over a time period of $t_n - t$, where $t_n - t$ repre-sents a large positive integer, the more consistent the model. Mathematically, can be represented as Equation (7):

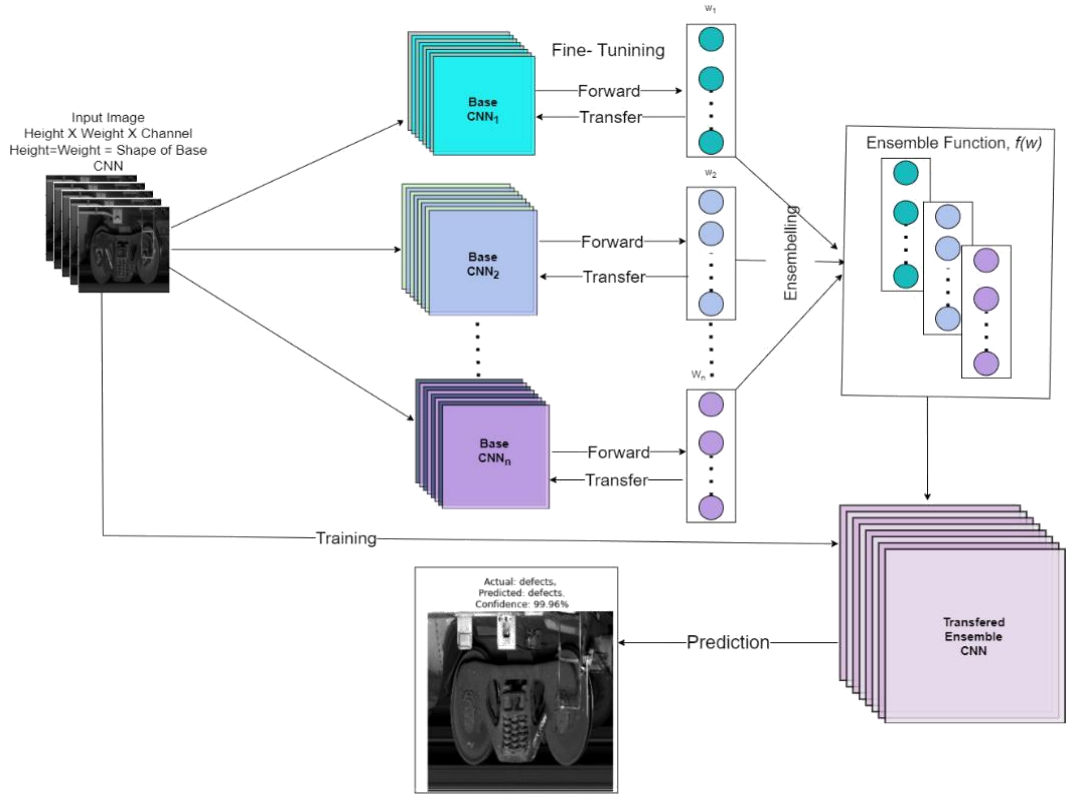

Fig. 1: System Architecture of Reusable AI-Enabled Defect Detection

$$\lim_{t_n - t \to \infty} \epsilon = 0 \quad (7)$$

This indicates that as the time period $t_n - t$ approaches infinity, the value of $\epsilon$ approaches 0, indicating a higher level of consistency in the model's performance. In other words, the smaller the value of $\epsilon$ and the longer the time period considered, the more consistency is demonstrated by the model. By solving this optimization problem, we can obtain a reusable AI model that not only optimizes the loss but also maintains consistent performance on the validation set, indicating its robustness and reliability.

## 3.2 Overview of System Architecture

Let us consider a scenario for a railway inspection system that captures images of rails and wheels. To identify defects in these components, multiple CNNs can be trained on a dataset of rail and wheel defect images, which can create a set of pre-trained CNN models. To ensure diversity among the models, each CNN can select training sets randomly from the dataset using different hyperparameters, architectures, or data pre-processing techniques. Finally, the resulting models can be reused in a transfer-ensembled approach. The diversity among the pre-trained models ensures that the transfer-ensembled model can achieve consistent performance across various scenarios. Therefore, the proposed model can serve as a reusable tool for defect classification tasks in railway inspection systems. The proposed system architecture for the reusable AI-enabled defect detection is illustrated in Figure 1. A brief overview of the architecture is given following.

- The architecture includes fine-tuned models fine-tuned CNN models (CN $N_1$, CN $N_2$, ·CN $N_n$) , providing transferred knowledge ($w_1$, $w_2$, ·$w_n$) of exist-ing neural network models. In this study, we employed VGG19, RESNET50, and MobileNetV2.
- Then the models are then concatenated by ensembling w = ($w_1$, $w_2$, ·$w_n$) using an ensembling function f(w).
- The ensembling function of the predictions with the combined best accuracy at the lowest loss. For example, if three different NN models classify images with 80%, 84%, and 86% accuracy respectively, then an ensemble of these three models can provide, for example, 88% accuracy. The selection criteria are usually weighted or an unweighted average of accuracy.

The type of defects varies in size, shape, and texture; and the ensemble technique provides similar accuracy by combining the probability score predicted by multi-ple NN. The detail process of the architecture is detailed in the following section. The methodology of the proposed system architecture is elaborated in the following sections.

## 3.3 Fine-Tuning CNN Parameters

We adopted the fine-tuning approach for extracting parameters of the pre-trained CNN models with rail defects. For example, VGG-19 is a CNN of 19 layers and trained with a huge amount of colored image samples from the Imagenet database. The layers of a pre-trained network are then modified to produce fine-tuned network for re-training the network with new training samples and labels. So that, the network can be re-trained with new training samples and labels. This approach of transferring learned parameters can mitigate the challenge of training a CNN from scratch with few samples. Because to train a CNN model from the beginning- an input image is passed through a series of convolutional and pooling layers to recognize image features and reduce the dimensionality of the image array. -

The small training sample size for newly added defect types overfit a fresh CNN model. A good illustration of such a case is the Canadian CPR Dataset. Because CPR dataset includes numerous data types but there is not sufficient data per defect type to train a multi-defect classifier. The fine-tuned model was therefore adopted to train a new network with pre-trained weights of an existing high-performance model. In Figure 2, the process of fine-tuning an existing network to reuse it for a new dataset has been illustrated.

1. Remove the input layer of CN $N_i$ and add a convolutional layer $Conv_{ti}$ for the defect dataset.

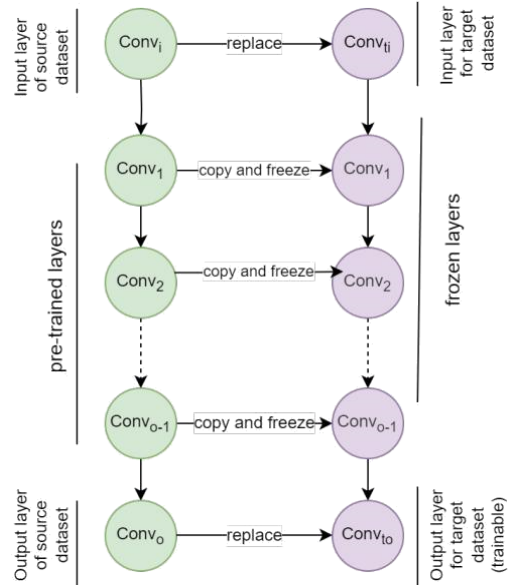

Fig. 2: Process of fine-tuning neural network to transfer knowledge from an existing network to a new network.

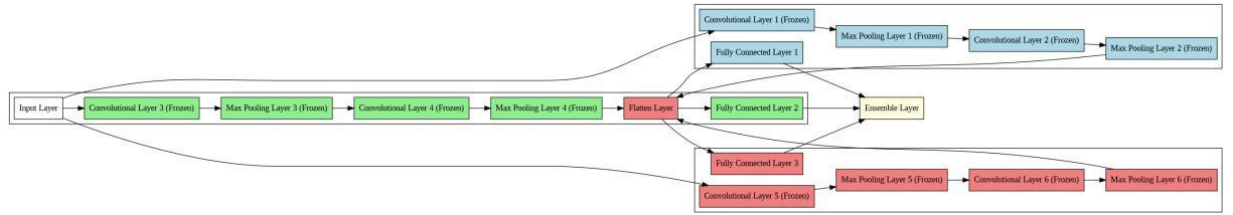

Fig. 3: Architecture of transfer learning ensemble CNN model

2. Copy and freeze all the convolutional layers until the output layers ($Conv_i$-$Conv_{o-1}$). These are the layers containing pre-trained weights w that we might need to extract through backpropagation if we were to train a model from scratch for a target dataset. The freezing of layers can be presented as Equation:

$$Conv_j^{(t+1)} = Conv_j^{(t)} \quad \text{for } j = i, i+1, \ldots, o-1 \qquad (9)$$

where $Conv_j^{(t)}$ denotes the j-th convolutional layer at training step t.
3. The convolutional layers are frozen in the transfer learned neural network to avoid the risk of losing the "good features" that have been adopted from the existing network by forward propagating the weights. The forward propagation through the fine-tuned layers can be represented as:

$$Conv_{to} = \sigma(Conv_{to-1} \cdot w_i) \qquad (10)$$

where $\sigma$ represents the activation function and $\cdot$ denotes the convolution operation.

4. Finally, remove the output layer of CN $N_i$ and add a new output layer $Conv_{to}$ as per the label (e.g., defect or normal) and number of outputs (e.g., 2) of the target defect dataset. The target output layer utilizes the learned weights w to learn patterns but is trained with new data. If n = 3, the visual representation of three pre-trained CNN models connected to an ensemble layer based on their lowest loss is illustrated in Figure 3

## 3.4 Ensembling Fine-Tuned CNNs

To address the challenge of balancing sample size and overfitting when training new deep networks with new data, we proposed lifetime modeling in a previous research study [35]. The idea behind lifetime modeling is that transfer knowledge can be used to classify new tasks. This comes into play when the model's deployment environment is altered, and the feature or data subspace changes from the source domain to the target domain.

---
**Algorithm 1** Algorithm for Ensembling Fine-tuned CNN Models
---
1: Let M be the set of fine-tuned models
2: Let N be the number of models to ensemble
3: Let $L_{min}$ be the minimum loss among the selected fine-tuned models
4: Let $f_{ens}$ be the ensemble model
5: Initialize $f_{ens}$ as an empty ensemble model
6: for i = 1 to N do
7:     Select a fine-tuned model $f_i$ from M
8:     if $f_i$ has the same shape as the defined shape then
9:         Add $f_i$ to $f_{ens}$
10:    end if
11: end for
12: Calculate the loss for each $f_i \in f_{ens}$
13: $L_{min} \leftarrow \min_{f_i \in f_{ens}} Loss(f_i)$
14:    Obtain the prediction y using $f_i$ with $L_{min}$: $y = f_i(x)$
15:    return y

---

Although the limited sample size issue is addressed with this transfer learning approach, the randomly varied features (e.g., size, shape, and texture) of data (e.g., image of defective parts) pose the bottleneck issue. Precisely, depending on a single backbone network has a risk of significant loss and lower performance, in case the selected pre-trained network is unable to handle samples having randomly changed features. Therefore in this study, we propose the ensembling of multiple transfer learned models to reduce the risk of total failure with a single network. The procedure of ensembling fine-tuned models is illustrated in Algorithm 1. The steps for ensembling the fine-tuned models are following.

- A fine tune model is selected and the shape of this model is compared to the defined shape to determine whether the fine-tuned model can be included in the list of selected backbone models or not.
- The model with the same shape is added to the ensemble model.
- Step 1-2 is repeated for a finite number of times n; where n is the number of models we want to ensemble.
- Loss for each of the selected fine-tuned models is obtained.
- The prediction by the model with minimum loss is selected as the final decision by the transfer learned ensemble model.

### 3.5 Complexity Analysis

The time complexity of Algorithm 1 is influenced by various factors, including the number of fine-tuned models (M), the complexity of the loss function (L), the size of the input data (D), the number of selected models (N), the complexity of the prediction function (P), and the size(e.g., shape) of the input (I). Considering the above steps, the overall time complexity of the algorithm can be approximated as Equation (11).

$$O(1) + O(M) + O(L * D) + O(N) + O(P * I) \quad (11)$$

By simplifying, we can represent the time complexity as Equation (12).

$$O(M + L*D + N + P*I) \quad (12)$$

To explain the complexity, the initialization step of the algorithm has a constant time complexity, which is theoretically considered efficient. The model selection step has a linear time complexity of $O(M)$, where M represents the number of fine-tuned models. If M is finite and optimally small, the impact on the overall complexity is minimal. Therefore, the number of models should be selected optimally. Similar complexity is observed for the determination of the minimum loss among the selected models as it also has linear time complexity of $O(N)$.

## 4 Data

We trained the model using the Canadian Pacific Railway (CPR) dataset to train the defect classifier. The dataset contains 3000 images of railway parts, among which 1000 images belong to the defect group, and 2000 images belong to the normal group. The dataset includes different views of the images, including the top view, bottom view, and a side view to better define the defective status of the parts.

The raw dataset has 404 unique labels, which means we do not have enough images per label. So, we sampled the data based on the parts label. For instance, Broken Flange WM66- L3 Verify, Broken Flange WM66- L4 Monitor, Broken Flange WM66-L4 Verify, Broken Flange WM66- R1, Broken Flange WM66- R1 Monitor, Broken Flange WM66- R1 Verify, Broken Flange WM66- R2 Verify, Broken Flange WM66-R2 Verify, etc. were all considered under Broken Flange defect group. However, the dataset was still significantly imbalanced as well as insufficient to train a DL model like CNN.

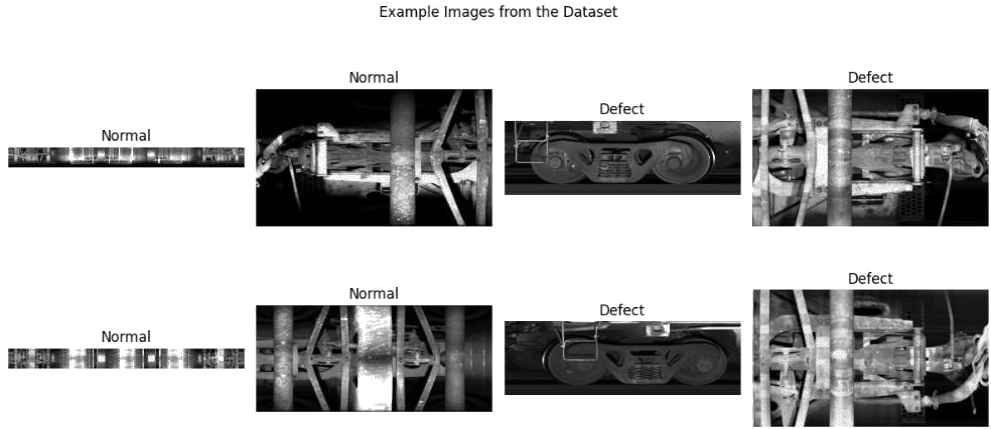

Fig. 4: Example Images from the CPR Dataset: This figure displays a mix of randomly chosen images, providing visual representation of the dataset used for the defect classification task.

In this study, we performed two-class classification with the CPR dataset to show the effectiveness of our proposed model in both cases. Some examples from the dataset are illustrated in Figure 4. For the binary classification, we considered the parent class defect and normal to mitigate the trade-off between sample size and class size. In this way, we got 1000 defect images and 2000 normal images. We split the dataset into 60% training-25% and 25% testing datasets following the state-of-the-art in the field of deep learning. The findings from the binary classification can also contribute to the subsequent multi-class classification.

## 5 Experiment and Result Analysis

In this section, we detail the data, empirical analysis, and comparative analysis of the proposed and existing models. The overview of the three different experiments conducted in this study is tabulated inTable 2. Here, training parameters have been chosen after several trial and error attempts.

We employed a powerful configuration to ensure efficient and accurate computations while training the CNN models. The processing power was provided by an Intel® Xeon® W-2133 CPU clocked at 3.60 GHz, allowing for swift and reliable data process-ing. To handle complex graphics-related tasks, we utilized the NVIDIA GeForce RTX 2080 GPU, which deliveres high-performance computing for deep learning algorithms. The machine was equipped with a substantial 64 GB of RAM, providing sufficient memory to handle complex model architectures. Running on a 64-bit operating sys-tem, specifically Windows 10 Enterprise edition, our system benefited from the latest software advancements and optimizations.

Table 2: Overview of the experiments

| Experiment | Experiment 1 | Experiment 2 | Experiment 3 |
|---|---|---|---|
| Objective | Augmentation for rail defect classification | Transfer learning model for rail defect classification | Ensemble model for rail defect classification |
| Model Architecture | CNN with sequential augmentation layer | Fine-tuned VGG-19, MobileNetV3, and RESNET50 | Concatenation of three fine-tuned models |
| Preprocessing | Resized to consistent shape and rescaled, Zoomed, Rotated | Resized to 224x224 | Network size was defined to 224x224 |
| Trainable Layers | 183,682 parameters | Frozen layers till dense layer | Frozen fine-tuned layers |
| Training Parameters | Batch size: 32, Epochs: 50 | Batch size: 32, Epochs: 10 | Learning rate scheduler: 0.003, Epochs: 20 |
| Performance Metrics | Accuracy, Loss, Batch Prediction | Accuracy, Loss | Accuracy, Loss, Precision, Recall |

To classify a defected and non-defected images, the predictions of the fine-tuned models are combined using the sigmoid function $f(\cdot)$ to obtain the probability for the positive class:

$$\hat{y} = f(w_1, w_2, \ldots, w_n) \quad (13)$$

where $\hat{y}$ represents the predicted probability for the positive class, and $w_1, w_2, \ldots, w_n$ are the weights of the fine-tuned models.

After obtaining the predicted probability, a threshold is applied to determine the final class label. If the predicted probability is above the threshold, the input image is classified as the positive class (defect); otherwise, it is classified as the negative class (normal).

During the training process, the parameters $w_i$ of each model are optimized to minimize the binary cross-entropy loss function $L_b$:

$$w_i^* = \arg\min_{w_i} L_b(D, w_i) \quad (14)$$

where $L_b$ calculates the binary cross-entropy loss between the predicted probabilities and the ground truth labels.

## 5.1 Evaluation of Data Augmented CNN Model

Data augmentation is considered a potential approach to address data imbalance issues. To evaluate the impact of using an augmented CNN model, we initially applied a data augmentation layer to a baseline CNN model and compared the performance. In this section, we present the experiment and outcomes of this evaluation.

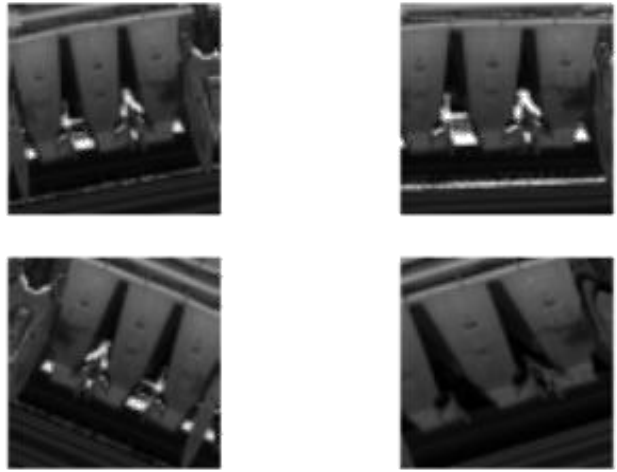

Fig. 5: Randomly augmented images of rail parts for enhancing dataset

### 5.1.1 Experiment Setup

In the previous section, it was noted that the raw dataset used in this study consisted of 81 unique classes, which can pose a challenge in developing accurate predictive models for railway assets due to their inherent heterogeneity. Manual feature extraction or data preprocessing can be time-consuming and may not be practical for large datasets. To address this challenge, a CNN classifier with a sequential augmentation layer was implemented. The augmentation layer randomly zooms in and out of images and rotates them at different angles to augment the data and mitigate the issues related to insufficient data.

An example of the augmentation layer task is illustrated in Figure 5. To standardize the pixel value, the layer resizes the images to a consistent shape at 255 and then re-scales them at (1./255). To further augment the data, horizontal and vertical random flips and random rotation at 0.2 were applied. Conventional layers with a 3 x 3 spatial size kernel, a stride size of 1, and a padding of 2 were used. Max pooling with a kernel size of 2x2 and zero padding was applied. As the dataset had two target classes, the Rectified Linear Unit (Relu) activation function was used for normalization. The model had 183,682 trainable parameters, which were optimized through several rounds of parameter tuning. A batch size of 32 and 50 epochs was used for training, with a split of 80:10:10 for training, testing, and validation data, respectively.

### 5.1.2 Findings

To evaluate the impact of adding the augmented layer, we compared the performance of the model with and without the augmented layer. It is evident from Figure 7, that the model with augmented layer classified defect and normal image with high confidence score. As shown in Figure 6, the training accuracy improved by nearly 20% after adding the augmented layer. Similarly, the validation accuracy increased from 80% to 90% and both training and validation loss decreased. These improvements suggest that the model may have been suffering from data overfitting issues before the addition of the augmented layer.

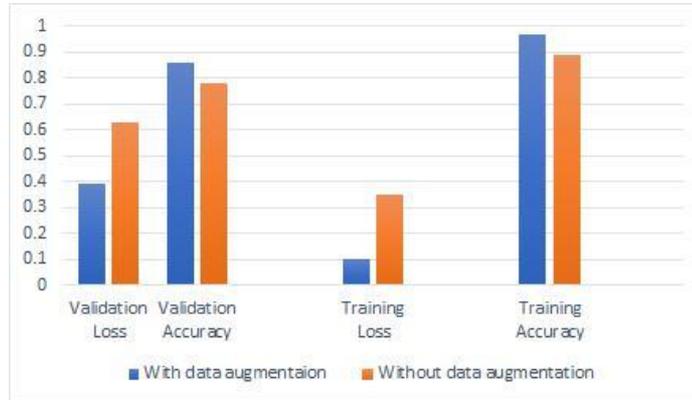

Fig. 6: Comparison of performance with and without augmentation layer

Overall, the prediction confidence score was good, as Figure 8 demonstrates that the validation loss for testing started to drop after five epochs and fluctuated through-out the epochs. We investigated the reason for the jump in the validation loss at epoch 40 and found that some of the defect and normal images were visually identical (see Figure 9), making it challenging to differentiate them even manually. We obtained the feature of these two images and found that the feature pixels quite resemble each other. Furthermore, By plotting the correlation matrix as a heatmap in Figure 10, we can observe patterns and relationships among the features of two resembled defective and normal images. Here, areas of high correlation (brighter colors) indicate a strong correlation between those features. On the contrary, areas of low correlation (darker colors) suggest a weaker relationship or independence between the features. It is evi-dent from the heatmaps of the images that they have similar feature patterns. This implies that the regions of the images share common patterns or characteristics for the defect and normal images.

In summary, although the data augmentation improved the model's accuracy, it was not able to handle the fluctuations in performance due to the overfitting issue. Therefore, we applied fine-tuning in the next phase to further improve the model's performance.

### 5.2 Evaluation of Fine-Tuned Model

The classification performance of different models on a training and validation set was compared in this study. Transfer learning was utilized as a method to enhance the performance of the models. In this section, we present the experiment details of the evaluation.

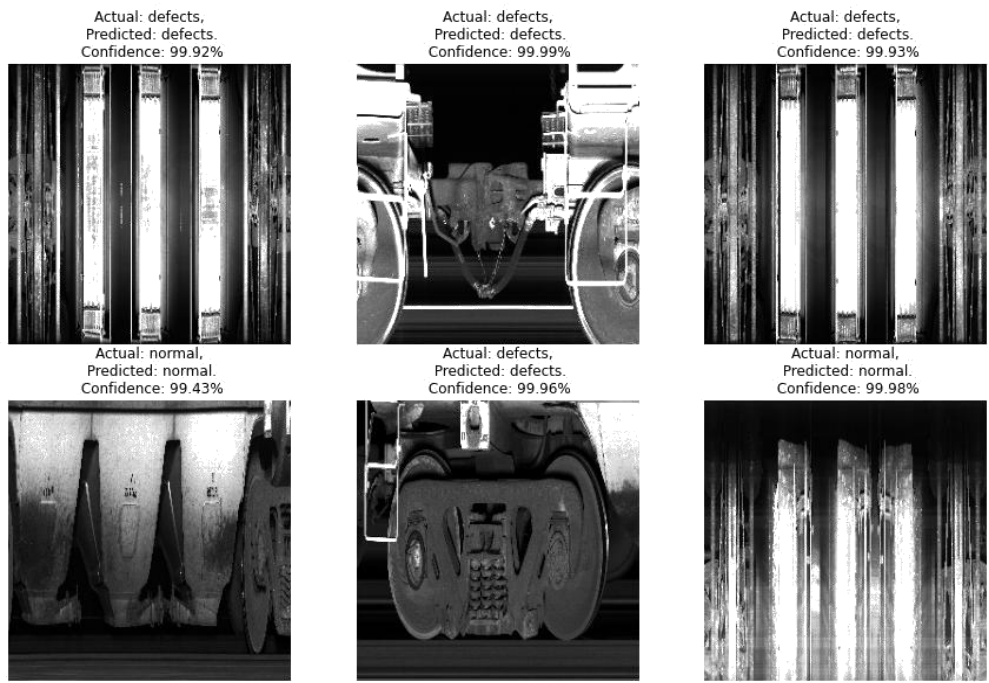

Fig. 7: Confidence score of defect detection on a batch of test image

### 5.2.1 Experiment Setup

The models are pre-trained on ImageNet database and accept 224X224 images as input. Therefore, we resized the training images to fulfill this requirement. We freeze the layers till the dense layer by setting the trainable layer parameter to false. Instead of freezing all layers, a few layers could be trained. However, due to the moderate sample size, it was tedious to determine the exact layers to be trained.

### 5.2.2 Findings

Figure 11 shows the comparison of the classification performance of individual models on the training and validation sets. The results demonstrated in Figure 8, shows that the fine-tuned models achieved higher accuracy and lower loss earlier than the augmented CNN model which required approximately 40 epochs to reach similar levels of performance.

Moreover, the fine-tuned models exhibited reduced fluctuation after the initial epochs, and the gap between validation and training performance decreased. Specifi-cally, for the RESNET50 model, there was a reduction in fluctuation and an increase in accuracy after epoch 6.

At some points, the training and validation curves intersected, indicating successful mitigation of the overfitting issue at those epochs. This suggests that transfer learning can be an effective approach for improving the classification performance of the models as well as improving the stability of the models during training.

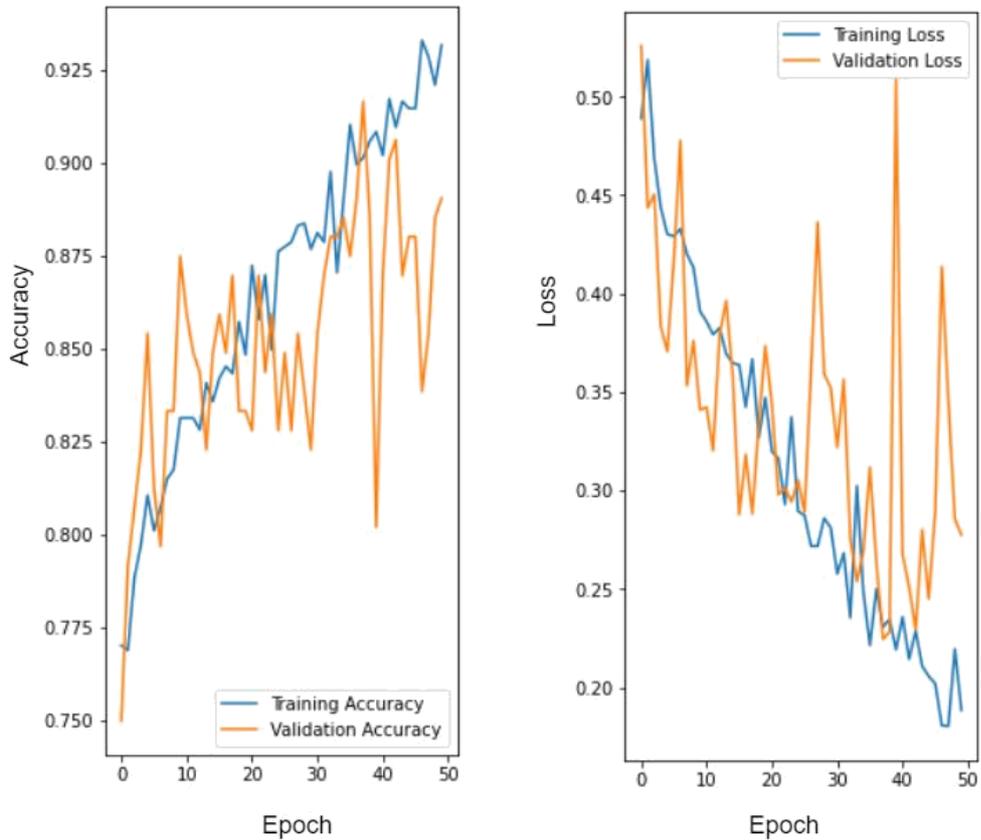

Fig. 8: Accuracy and loss graph after adding Data Augmentation

However, the performance evaluation of different fine-tuned models on the same data in this study revealed a variation while training. For VGG, the minimum vali-dation loss of 0.0067 was achieved during the first epoch, while for MobileNet, it was obtained during the third epoch with a value of 0.0055, and for ResNet, it was reached during the sixth epoch with a validation loss of 0.0103. Interestingly, the validation accuracy for VGG was 0.92 during the minimum validation loss, while for MobileNet it was 0.91, and for ResNet, it was 0.94.

These results indicate that while the ResNet model achieved the highest validation accuracy, the MobileNet model showed the least loss with the lowest accuracy. As a result, it is challenging to select a single model that performs the best. The findings of this experiment underscore the importance of careful model selection when developing DL models. To address this requirement, we proposed an ensemble model, which aims to obtain the best potential models from a single model. The performance of the transfer-learned ensemble model is discussed in the next section.

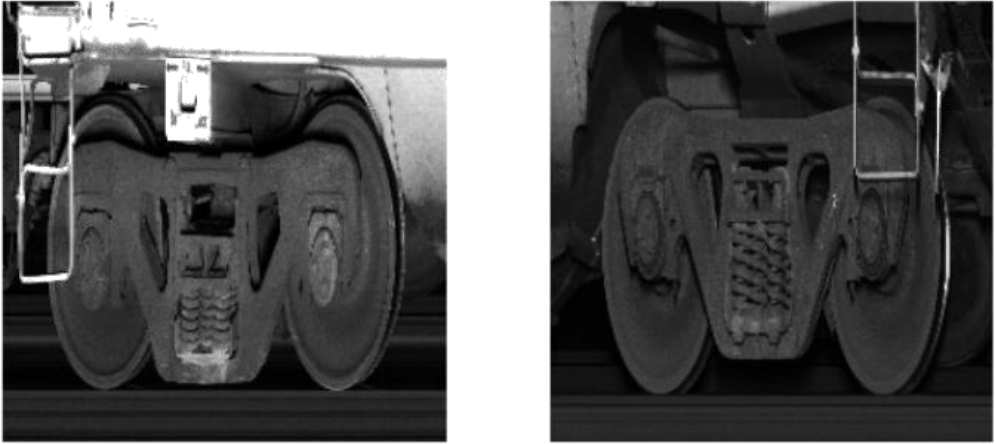

Fig. 9: Identical defect and normal image causing bounce in prediction accuracy. Left one: defect, Right one: normal

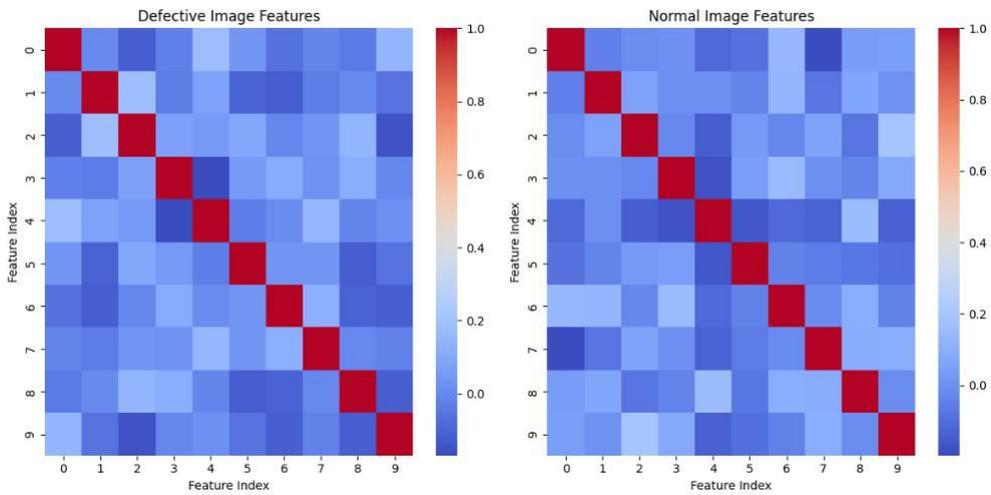

Fig. 10: Heatmap of identical samples. Left one: defect, Right one: normal. Both of the images were resized into 224X224 before extracting the heatmap. The Feature Index represents the position of the features in the correlation matrix.

## 5.3 Evaluation of Proposed model

In order to obtain the proposed reusable model, we combined the previously fine-tuned VGG19, RESNET50, and MobileNetV3 CNN models into an ensemble model. This section provides a detailed account of the experiment setup and findings of the evaluation.

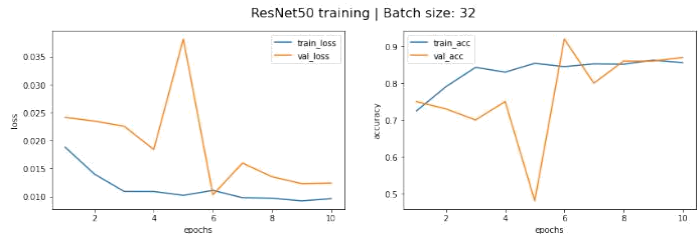

(a) Loss and Accuracy graph of RESNET50

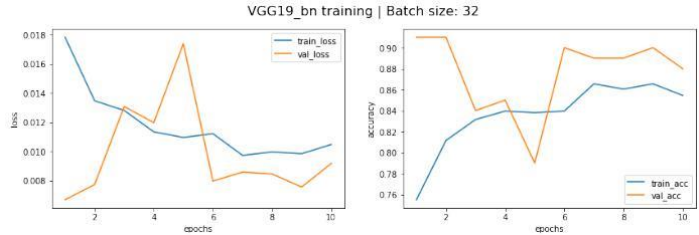

(b) Loss and Accuracy graph of VGG-19

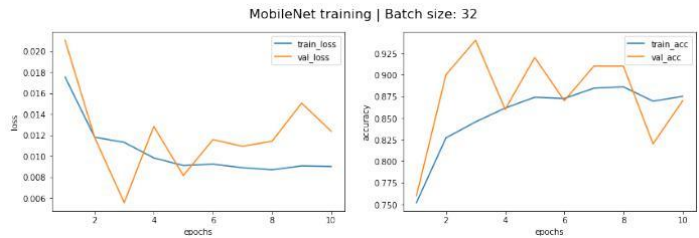

(c) Loss and Accuracy graph of MobileNetV3

Fig. 11: The accuracy and loss curves of three fine-tuned networks on the CPR train-ing and validation sets. X-axis represents epochs, and Y-axis represents the accuracy. train ACC = accuracy curve of the training set, val ACC = accuracy curve of the validation set, train loss = loss curve of the training set, and val loss = loss curve of the validation set

### 5.3.1 Experiment Setup

To obtain the ensemble model, the three models were concatenated as linear models. For selecting prediction, the minimum loss was obtained. Since multiple fine-tuned models had good performance in different epochs, the number of epochs was increased to 20, and a learning-rate scheduler starting from 0.003 was used to mitigate the risk of overfitting.

### 5.3.2 Findings

The findings of the evaluation of loss and accuracy graphs are shown in Figure 12. From the graphs, it is evident that after the initial epochs, the loss and accuracy levels reached an acceptable point and remained constant at almost 0.0001 validation loss and 0.99 validation accuracy.

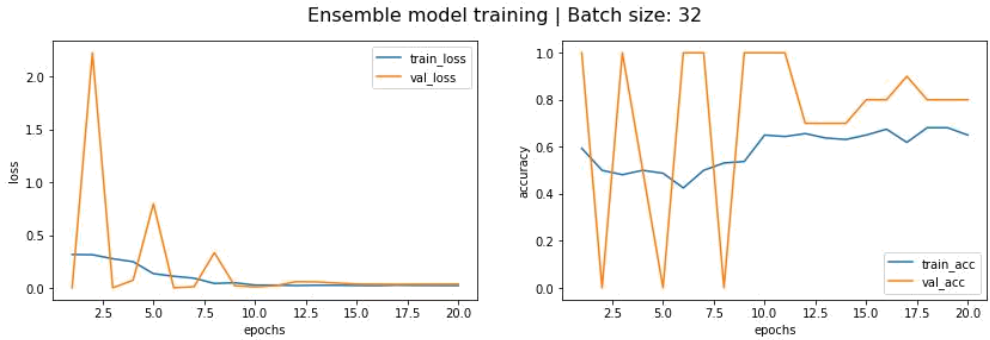

Fig. 12: Accuracy and Loss curve of proposed reusable model

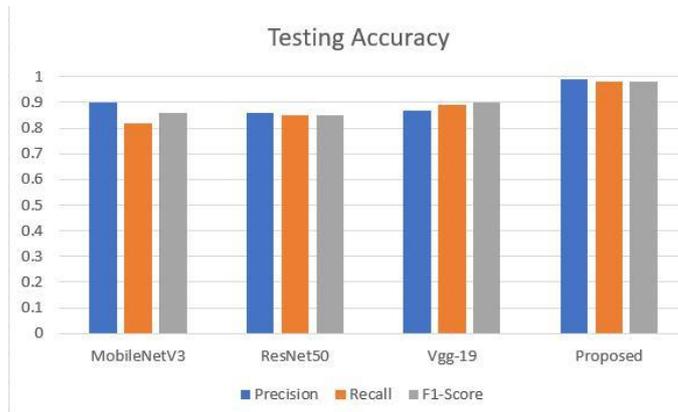

Fig. 13: Accuracy comparison between fine-tuned and proposed model

The bar diagram Figure 13, presents precision, recall, and F1-score for the individ-ual models, including MobileNetV3, ResNet50, VGG-19, and the proposed ensemble model. It is observed that the proposed ensemble model has the highest F1-score of 0.98, followed closely by MobileNetV3 and VGG-19 with F1-scores of 0.86 and 0.90 respectively. ResNet50 has the lowest F1-score of 0.85. The reason for the higher F1-score of the proposed ensemble model could be attributed to the fact that the ensemble model combines the strengths of the individual models and eliminates the weaknesses. Comparitavely, ResNet50 may have lower performance due to its architecture and complexity, which may have led to overfitting.

In a nutshell, the proposed ensemble model showed comparatively better results for the classification task, while the individual models also demonstrated reasonable performance. Subsequently, the challenge of achieving higher accuracy and lower loss from the same model was addressed through the ensembling of fine-tuned models. This approach effectively leverages the strengths of each model, resulting in a more reliable performance.

## 5.4 Comparison with Other Defect Classification Model

In this section, we present a comparison of the performance between our proposed model and other models used in various rail defect classification studies. While making the comparison, we encountered two challenges. Firstly, the majority of studies have trained their models using in-house and private datasets. One of the reasons behind this could be there is no publicly available benchmark dataset specifically designed for rail defect classification. Secondly, the selection of performance metrics has not been consistent across different studies. Most studies have focused solely on test accuracy and few have considered other essential metrics such as F1-score, precision, and recall. To address these challenges and provide we categorized the existing work into three groups based on the type of the models used. Table 3 summarizes the total data size, model, and obtained performance for rail defect classification.

Table 3: Comparison with existing work

| Study | Data Size | Model | Performance Obtained |
|---|---|---|---|
| [15] | 3,240 | Deep CNN (DCNN) | Test Accuracy: 92% |
| [16] | 4,220 | DCNN | Test Accuracy: 96.9% |
| [17] | 195 | YOLOv3 | Test Accuracy: 97% |
| [18] | No specific data size | 1D CNN | Test Accuracy: 97.8% |
| [19] | 317 | Deep CNN | Recall-80.48%, Precision-78.20%, F1-score-79.32% |
| [20] | 120 | YOLO, YOLO+MobileNet | Test Accuracy: 57.4%, 96.3% |
| [21] | 2,117 | Modified AlexNet, VGG | Test Accuracy: 90% |
| [22] | 400 | ResNet50 transfer learning | Test Accuracy: 94% |
| [23] | 138 | RESNET and DenseNET | Test Accuracy: 90% |
| [24] | 1000 | 1DCNN and LSTM | Dataset-2: Recall-0.9427, Precision-0.9176, F1-score 0.9300 |
| [25] | 147 | Modified YOLOV4 | Test Accuracy: 92.3%-96.3% |
| [27] | 244 | 10 CNN | Test Accuracy: 81.6%-91.89% |
| [36] | 189,88 | MobileNet + YOLO | Test Accuracy: 87.40% |
| Proposed | 3000 | RESNET50+VGG19+MobileNetV3 | Precision: 0.99, Recall: 0.98, F1-score: 0.98, Test-Accuracy:99% |

### 5.4.1 Comparison with Deep CNN Models

Faghih et al. [15] and Jamshidi et al. [16] employed Deep CNN (DCNN) models, achieving accuracies of 92% and 96.9%. While these models demonstrate the effective-ness of DCNNs in rail defect classification tasks, they may face limitations in dealing with complex and diverse defect types. Because if the sample size for training a new defect classifier is not enough, then a DCNN model can not be reused. Therefore, The performance could be improved by incorporating transfer learning techniques. In our proposed model we handled this and obtained better performance by leveraging pre-trained weights.

### 5.4.2 Comparison with Object Detection Models

Yanan et al. [17] utilized YOLOv3, while Liang et al.[20] employed YOLO and YOLO+MobileNet models. These models achieved accuracies ranging from 57.4% to 97%. The relatively lower accuracy of YOLO in [20] could be attributed to the limited dataset size (120 images). Although YOLO is a popular choice due to its accuracy in real-time object detection, it requires manually labeled bounding box samples to train the model. Authors may have mostly used lower sample size for training YOLO defect classifier, due to its labor-intensive and time-consuming data preparation step. Another reason could be the unavailability of the data. Surprisingly, the authors in [36] could obtain only 87% accuracy even though they employed a large dataset with 189,88 samples. This may happen due to the complexity of concatenating three bot-tleneck modules. By contrast, in our proposed model we performed the concatenation of the model based on the lowest loss. So that, the individual model complexity does not affect the prediction accuracy.

### 5.4.3 Transfer Learning Models

Kim et al. [21] utilized modified AlexNet and VGG models, achieving an accuracy of 90% on their dataset. Santur et al. [22] employed transfer learning with ResNet50, achieving an accuracy of 94%. Similarly, James et al. [23] utilized RESNET and DenseNET models, achieving an accuracy of 90%. Passos et al. [27] utilized 10 pre-trained CNN models and achieved accuracies ranging from 81.6% to 91.89% for defect classification. While these models perform reasonably well, ensembling techniques, such as combining predictions from multiple models or utilizing an ensemble of dif-ferent architectures, can help improve the overall performance by capturing diverse representations and reducing the impact of model biases.

Comparing to existing approaches in [19] and [22], our proposed model outperforms in terms of precision, recall, and F1-score. The higher precision at 0.99 indicates only 1% rate of misclassifying non-defective samples as defects. The higher recall at 0.98 demonstrates the model's capability to correctly identify actual defects, which is a crucial measure in the context of rail defect diagnosis. The F1-score is also at 0.98 demonstrating the overall measure of the classification performance by our proposed model. As most of the studies have trained the CNN models for in-house dataset the test accuracy is good (above 90%) for several studies. However, our proposed model not only demonstrated comparative better test accuracy (at 99%) but also showed decent performance for other accuracy measures. Moreover,we have obtained consistent validation accuracy and validation loss.

The good performance of our proposed model is obtained due to harnessing the capability of three CNN models- namely, RESNET50, VGG19, and MobileNetV3. To be specific, RESNET50 allows capturing of intricate features and patterns within the rail defect images. VGG19 enables the extraction of high-level features. MobileNetV3, on the other hand, is a lightweight and efficient architecture, facilitating faster infer-ence and ensuring higher accuracy. Therefore, if a new portion of samples does not fit a specific model, the other model can ensure consistency in the accuracy.

## 6 Limitations and Future Work

The proposed work has some limitations that need to be addressed. The model is evaluated by performing binary classification, which means that it can accurately distinguish between normal and defective parts. However, we have evaluated the per-formance of the model in terms of different accuracy measures and obtained high and consistent accuracy. Further research is needed to investigate the effectiveness of the model in identifying various types of defects or locating the precise defect location.

Although ensemble CNN combines multiple models to improve the accuracy of predictions, it can be computationally expensive and time-consuming. Visual transformers (ViT) [37], on the other hand, ViT is a type of DL model that has recently shown remarkable performance on various CV tasks, including image classification and object detection. They have the ability to capture global spatial information from an image and attend to specific regions of interest, making them highly effective for com-plex visual recognition tasks. Therefore, for further research, visual transformers, as a type of DL model, can be explored as an alternative approach to improve the speed along with the accuracy of the railway defect detection system.

# 7 Conclusion

In conclusion, we have presented an approach for railway defect detection combining transfer learning and an efficient reusable model. We demonstrated that the ensem-bling and fine-tuning of transfer learning with VGG-19, MobileNetV3, and ResNet-50 models achieve high accuracy at different epochs. Our proposed model provided 99% accuracy reduced overfitting, and exhibited stable performance after a certain epoch. Overall, our proposed approach contributes to the development of a more efficient and effective method for railway defect detection, which can potentially improve railway safety and reliability. Taking the linear complexity of model speed in consideration, the use of a visual transformer model can be a further extension of ensembled trans-fer learning to speed up training without sacrificing accuracy. Therefore, in future work, we aim to investigate the effectiveness of visual transformers as an alternative to ensemble learning.

# Declaration

## Competing Interests

The methods presented in this paper are highly demanded in developing AI-enabled digital twin for Industry, particularly for application domains that requires high scalability of the predictive models and adaptability to changing operational environments

## Authors contribution statement

- Rahatara Ferdousi: Conceptualization, Methodology, Writing - Original Draft Preparation, Visualization, Validation, Software.
- Fedwa Laamarti: Writing - Review & Editing, Visualization, Validation, Data Curation.
- Chunsheng Yang: Writing - Review & Editing, Supervision, Visualization, Validation, Project administration.
- Abdulmotaleb El Saddik: Writing - Review, Funding acquisition, Supervision.

### Ethical and informed consent for data used

We have signed a business agreement with the data provider. We have consent for publishing the result.

### Data availability and access

The datasets analyzed during the current study are not publicly available. Due to the terms of the business agreement, we cannot disclose the data for public access

## Acknowledgement

This research is supported in part by collaborative research funding from the National Program Office under the National Research Council of Canada's Artificial Intelligence for Logistics Program. The project ID is AI4L-123.

## References

[1] Gao, J., Lin, L., He, D., Wang, R.: Digital twin and ai techniques in railway transportation: a review. Journal of Ambient Intelligence and Humanized Computing, 1–15 (2021)

[2] Pan, J., Jiang, Z., Gao, X., Wu, D.: Railway track inspection and defect detection based on unmanned aerial vehicles: a review. Journal of Intelligent Transportation Systems 24(1), 1–16 (2020)

[3] Sun, W., Zeng, X., Huang, S., Wang, S.: Railway defect detection based on computer vision technology: a survey. Journal of Intelligent Transportation Systems 22(1), 43–55 (2018)

[4] Zhao, C., Wang, R., Huang, Z., Li, H., Zhang, Z., Sun, Y.: Deep learning-based automatic recognition of rail surface defects using a convolutional neural network. Sensors 17(10), 2384 (2017)

[5] Ferrer, B., Kerckhof, R., Bleys, B., Demeulemeester, E.: Digital twins in industry: A survey. Computers in Industry 109, 3–19 (2019)

[6] Meng, Y., Xu, H., Ma, Z., Zhou, J., Hui, D.: Detail-semantic guide network based on spatial attention for surface defect detection with fewer samples. Applied Intelligence 53(6), 7022–7040 (2023)

[7] Yosinski, J., Clune, J., Bengio, Y., Lipson, H.: How transferable are features in deep neural networks? In: Advances in Neural Information Processing Systems, pp. 3320–3328 (2014)

[8] Ferdousi, R., Laamarti, F., Yang, C., El Saddik, A.: Railtwin: A digital twin framework for railway. In: 2022 IEEE 18th International Conference on Automa-tion Science and Engineering (CASE), pp. 1767–1772 (2022). IEEE

[9] Wang, X., Shen, C., Xia, M., Wang, D., Zhu, J., Zhu, Z.: Multi-scale deep intra-class transfer learning for bearing fault diagnosis. Reliability Engineering & System Safety 202, 107050 (2020)

[10] Rodrigues, P.C.L.: Detecção de anomalias em trilho utilizando visão computacional. Master's thesis, Instituto Federal do Espírito Santo, Programa de Pósgraduação em Tecnologias Sustentáveis (2020)

[11] Zhang, S., Wei, H.-L., Ding, J.: An effective zero-shot learning approach for intel-ligent fault detection using 1d cnn. Applied Intelligence 53(12), 16041–16058 (2023)

[12] Tan, M., Le, Q.V.: Efficientnet: Rethinking model scaling for convolutional neural networks. In: International Conference on Machine Learning, pp. 6105–6114 (2019). PMLR

[13] Zhang, Y., Wang, Y., Zhao, Q., Gao, H.: A survey of recent advances in vehicle defect detection using computer vision. Measurement 143, 277–289 (2019)

[14] Zhang, H.-D., Yuan, X., Li, D.-Y., You, J., Liu, B., Zhao, X.-M., Cai, W.-M., Ju, S.: An effective framework using identification and image reconstruction algorithm for train component defect detection. Applied Intelligence 52(9), 10116–10134 (2022)

[15] Faghih-Roohi, S., Hajizadeh, S., Núñez, A., Babuska, R., De Schutter, B.: Deep convolutional neural networks for detection of rail surface defects. In: 2016 International Joint Conference on Neural Networks (IJCNN), pp. 2584–2589 (2016). IEEE

[16] Jamshidi, A., Faghih-Roohi, S., Hajizadeh, S., Núñez, A., Babuska, R., Dollevoet, R., Li, Z., De Schutter, B.: A big data analysis approach for rail failure risk assessment. Risk Analysis 37, 1495–1507 (2017)

[17] Yanan, S., Hui, Z., Li, L., Hang, Z.: Rail surface defect detection method based on yolov3 deep learning networks. In: 2018 Chinese Automation Congress (CAC), pp. 1563–1568 (2018). IEEE

[18] Zhandong, Y., Shengyang, Z., Xuancheng, Y., Wanming, Z.: Vibration-based damage detection of rail fastener clip using convolutional neural network: Experiment and simulation. Engineering Failure Analysis 119 (2021)

[19] Xiao, L., Wu, B., Hu, Y., Liu, J.: A hierarchical features-based model for freight train defect inspection. IEEE Sensors Journal 20(5), 2671–2678 (2019)

[20] Liang, Z., Zhang, H., Liu, L., He, Z., Zheng, K.: Defect detection of rail surface with deep convolutional neural networks. In: 2018 13th World Congress on Intelligent Control and Automation (WCICA), pp. 1317–1322 (2018). IEEE

[21] Kim, H., Lee, S., Han, S.: Railroad surface defect segmentation using a modified fully convolutional network. KSII Transactions on Internet & Information Systems 14 (2020)

[22] Santur, Y., Karak"ose, M., Akın, E.: Learning based experimental approach for condition monitoring using laser cameras in railway tracks. International Journal of Applied Mathematics Electronics and Computers (Special Issue-1), 1–5 (2016)

[23] James, A., Jie, W., Xulei, Y., Chenghao, Y., Ngan, N.B., Yuxin, L., Yi, S., Chandrasekhar, V., Zeng, Z.: Tracknet-a deep learning based fault detection for railway track inspection. In: 2018 International Conference on Intelligent Rail Transportation (ICIRT), pp. 1–5 (2018). IEEE

[24] Zhang, Y., Cao, L., Wang, J., Yu, Y.: A multi-class defect detection algorithm for rail surface based on inception-v3. Journal of Physics: Conference Series 1216(5), 052018 (2019)

[25] Anwar, N., Shen, Z., Wei, Q., Xiong, G., Ye, P., Li, Z., Lv, Y., Zhao, H.: Yolov4 based deep learning algorithm for defects detection and classification of rail sur-faces. In: 2021 IEEE International Intelligent Transportation Systems Conference (ITSC), pp. 1616–1620 (2021). IEEE

[26] Chen, R., Jin, C., Zhang, Y., Dai, J., Lv, X.: Digital twin for equipment management of intelligent railway station. In: 2021 IEEE 1st International Conference on Digital Twins and Parallel Intelligence (DTPI), pp. 374–377 (2021). IEEE

[27] Passos, R.A.d.S.L., Ferreira, M.P., Silva, B.-H.d.A., Lopes, L.A.S., Ribeiro, H., Santos, R.P.: An in-depth assessment of convolutional neural networks for rail surface defect detection. Research, Society and Development 11(8), 12211830252– 12211830252 (2022)

[28] Afshari, S.S., Enayatollahi, F., Xu, X., Liang, X.: Machine learning-based methods in structural reliability analysis: A review. Reliability Engineering & System Safety 219, 108223 (2022)

[29] Xu, Y., Wang, H., Lu, Q., Liu, Y.: Rail surface defect detection based on mobilenet. Journal of Physics: Conference Series 1589(5), 052058 (2020)

[30] Xiaoyu, T., Jinbo, H., Jiewen, F., Xihe, C.: Image segmentation and defect detection of insulators based on u-net and yolov4. () 52(6), 15–21 (2020)

[31] Xia, M., Shao, H., Williams, D., Lu, S., Shu, L., Silva, C.W.: Intelligent fault diag-nosis of machinery using digital twin-assisted deep transfer learning. Reliability Engineering & System Safety 215, 107938 (2021)

[32] Ariyachandra, M., Brilakis, I.: Digital twinning of railway overhead line equipment from airborne lidar data (2020)

[33] Li, H., Wang, F., Liu, J., Song, H., Hou, Z., Dai, P.: Ensemble model for rail surface defects detection. PLoS one 17(5), 0268518 (2022)

[34] Zou, Q., Chen, S.: Resilience-based recovery scheduling of transportation network in mixed traffic environment: a deep-ensemble-assisted active learning approach. Reliability Engineering & System Safety 215, 107800 (2021)

[35] Yang, C., Ferdousi, R., El Saddik, A., Li, Y., Liu, Z., Liao, M.: Lifetime learning-enabled modelling framework for digital twin. In: 2022 IEEE 18th International Conference on Automation Science and Engineering (CASE), pp. 1761–1766 (2022). IEEE

[36] Yuan, H., Chen, H., Liu, S., Lin, J., Luo, X.: A deep convolutional neural network for detection of rail surface defect. In: 2019 IEEE Vehicle Power and Propulsion Conference (VPPC), pp. 1–4 (2019). IEEE

[37] Carion, N., Massa, F., Kirillov, R., Girshick, R.: End-to-end object detection with transformers. European Conference on Computer Vision, 213–229 (2020)